\documentclass{Interspeech2024}

\interspeechcameraready 

\usepackage{cleveref}
\usepackage{cite}
\usepackage{subfig}
\usepackage{graphicx}
\usepackage{xcolor}

\newcommand{\rand}{{\it random}}
\newcommand{\homophone}{{\it near-homophones}}
\newcommand{\syn}{{\it synonyms}}
\newcommand{\same}{{\it same-words}}

\newcommand{\wavtovecB}{{\it wav2vec2.0-Base}}
\newcommand{\wavtovecL}{{\it wav2vec2.0-Large}}

\newcommand{\hubertL}{{\it HuBERT-Large}}
\newcommand{\hubertB}{{\it HuBERT-Base}}
\newcommand{\xlsr}{{\it XLS-R-300M}}

\newcommand{\wavlmL}{{\it WavLM-Large}}

\newif\ifdraft
\drafttrue

\definecolor{dkgreen}{RGB}{0,179,36}
\definecolor{dkred}{RGB}{240,0,0}
\definecolor{dkblue}{RGB}{0,100,240}
\definecolor{dkorange}{RGB}{230,115,0}
\definecolor{dkbrown}{RGB}{170,50,0}
\definecolor{chestnut}{rgb}{0.8, 0.36, 0.36}
\definecolor{pink}{RGB}{255,0,247}
\definecolor{amber}{rgb}{1.0, 0.75, 0.0}
\definecolor{amethyst}{rgb}{0.6, 0.4, 0.8}

\title{Self-Supervised Speech Representations are More Phonetic than Semantic}

\name[affiliation={1}]{Kwanghee}{Choi}
\name[affiliation={2}]{Ankita}{Pasad}
\name[affiliation={3}]{Tomohiko}{Nakamura}
\name[affiliation={3}]{Satoru}{Fukayama}
\name[affiliation={2}]{\\Karen}{Livescu}
\name[affiliation={1}]{Shinji}{Watanabe}

\address{
  $^1$Carnegie Mellon University
  $^2$Toyota Technological Institute at Chicago\\
  $^3$National Institute of Advanced Industrial Science and Technology (AIST)}
\email{\{kwanghec,swatanab\}@andrew.cmu.edu, \{ankitap,klivescu\}@ttic.edu, \{tomohiko-nakamura,s.fukayama\}@aist.go.jp}


\begin{document}

\maketitle

\begin{abstract}
Self-supervised speech models (S3Ms) have become an effective backbone for speech applications. Various analyses suggest that S3Ms encode linguistic properties. In this work, we seek a more fine-grained analysis of the word-level linguistic properties encoded in S3Ms. Specifically, we curate a novel dataset of near homophone (phonetically similar) and synonym (semantically similar) word pairs and measure the similarities between S3M word representation pairs. Our study reveals that S3M representations consistently and significantly exhibit more phonetic than semantic similarity. Further, we question whether widely used intent classification datasets such as Fluent Speech Commands and Snips Smartlights are adequate for measuring semantic abilities. Our simple baseline, using only the word identity, surpasses S3M-based models. This corroborates our findings and suggests that high scores on these datasets do not necessarily guarantee the presence of semantic content.
\end{abstract}
\section{Introduction}
Self-supervised speech models (S3Ms) have become the de facto standard for solving various downstream speech tasks \cite{baevski2020wav2vec,hsu2021hubert,chen2022wavlm,yang21c_interspeech,shi2023ml}.
Analysis studies suggest that S3Ms encode various linguistic properties related to: phonetics \cite{pasad2021layer,wells22_interspeech,abdullah23_interspeech,choi2022opening,choi2023understanding}, phonology \cite{martin23_interspeech}, morphology \cite{pasad2021layer,pasad2023comparative,pasad2023self}, syntax \cite{shen23_interspeech,pasad2021layer,pasad2023comparative,pasad2023self,DBLP:conf/emnlp/MohebbiCZA23}, and semantics \cite{ashihara23_interspeech,pasad2021layer,pasad2023comparative,pasad2023self}.

However, most existing literature either focuses only on phonetic analysis \cite{wells22_interspeech,choi2022opening,abdullah23_interspeech} or studies the existence of various linguistic properties \cite{pasad2021layer,pasad2023comparative,shen23_interspeech}.
We still do not understand if the S3M representations are better at encoding phonetic or semantic properties, or are equally good at both.
To answer that, we design a fine-grained analysis framework to study the word-level representations.
Specifically, we measure the similarity between word pairs, namely, \homophone \ and \syn.
For example, consider the two word pairs: (dog, dig) and (dog, puppy). 
While ``puppy" is semantically similar to ``dog," ``dig" is more phonetically similar to ``dog" while bearing no semantic similarity.

Our experimental results (\Cref{sec:analysis}) show that while semantically similar word pairs are closer than random word pairs, phonetically similar word pairs are significantly closer across all layers.
This behavior is repeatedly observed in various S3Ms, differing in pre-training objective, model size, and the language of the pre-training data.
Next, we seek to verify this implication on downstream task performance.
Previous studies show that S3Ms perform exceptionally well in semantic tasks such as intent classification (IC) \cite{yang21c_interspeech,arora21_interspeech,lugosch19_interspeech}.
For instance, the frozen S3M model accuracy for the Fluent Speech Commands (FSC) dataset \cite{qian2021speech} is over 99\% \cite{yang21c_interspeech}, making the task seem already solved.
However, we question whether these high scores are indicative of word meaning encoded within S3M representations.
To verify, we develop a simple baseline that uses only the word identity information.
A simple decision tree classifier trained on word identity performs better than frozen S3M representations on two IC datasets.
Thus, we conclude that S3M's impressive performance on these IC datasets does not indicate their semantic capabilities.

To summarize, in this work, we (i) contribute a carefully curated dataset of word pairs that distinguishes meaning and pronunciation (\Cref{sec:setup}), (ii) study various S3Ms and find that word-level S3M representations are more phonetic than semantic (\Cref{sec:analysis}), and (iii) critically study widely used intent classification tasks and conclude that the S3M performance on these tasks is not necessarily indicative of semantic knowledge (\Cref{sec:baseline}). 

\section{Experimental Setup}\label{sec:setup}
\subsection{Extracting synonyms and near homophones}\label{subsec:syn_hom}
To extract synonyms, we use WordNet \cite{fellbaum2010wordnet}, an English lexical database.
For simplicity, we regard all the words in synsets (cognitive synonyms) as synonyms.
In addition, to extract synonyms in different languages, we also use Open Multilingual Wordnet v1.4 (OMW) \cite{bond2012survey}.
We use NLTK \cite{DBLP:conf/acl/Bird06} to access both WordNet and OMW.
For example, there are 16 synsets of the word ``study''.
Examples are synset \texttt{learn.v.04} with words \{learn, study, read, take\}, or synset \texttt{report.n.01} with \{report, study, written report\}.

For near homophones, we first phonemicize each word using the CMU pronouncing dictionary \cite{weide1998carnegie} for English and Epitran \cite{mortensen2018epitran} for other languages.
Then we use the Levenshtein distance \cite{levenshtein1966binary,nerbonne1997measuring} to measure the phonetic distance between two phonemicized word pairs, divided by the longest of the two pairs to normalize the distance within the range 0 to 1.
Further, to yield a sensible threshold for defining the near homophones, we measure the phonetic distance between random word pairs in the LibriSpeech dataset.
We found that the top 0.1\% of the random word pairs have phonetic distance equal to or smaller than $0.4$.
Hence, we define near homophones as having normalized Levenshtein distance $d \leq 0.4$.
Example word pairs can be found in \Cref{tab:phonetic_dist}.

\begin{table}[t]
\caption{Example word pairs and their normalized Levenshtein distances.}
\label{tab:phonetic_dist}
\centering
\resizebox{0.45\textwidth}{!}{%
\begin{tabular}{l|l}
\toprule
Distance range     & Example word pairs and their distances                \\
\midrule
$0.0 < d \leq 0.2$   & [oysters, stirs (0.20)] [dripping, drilling (0.17)] \\
$0.2 < d \leq 0.4$ & [mind, mound (0.25)] [bread, braids (0.4)]            \\
$0.4 < d \leq 0.6$ & [socks, saxon (0.5)] [caffeine, patio (0.6)]          \\
$0.6 < d \leq 0.8$ & [morally, sausage (0.67)] [spring, constrain (0.75)]  \\
$0.8 < d \leq 1.0$ & [euclid, jack (0.83)] [cherry, shrank (1.0)]          \\
\bottomrule
\end{tabular}%
}

\end{table}

\subsection{Datasets}\label{subsec:datasets}
For English analysis, we use both dev-clean and test-clean subsets of the LibriSpeech dataset \cite{panayotov2015librispeech}, similar to \cite{pasad2023comparative,pasad2021layer}.
We also use existing LibriSpeech alignments \cite{lugosch19_interspeech} to extract timestamp information for each word.
Alignments are generated by the Montreal Forced Aligner (MFA) \cite{mcauliffe17_interspeech}.
To strengthen the statistical significance of our findings, we use statistical bootstrapping \cite{efron1994introduction}.
We randomly choose 10K word utterances and repeat the same experiment five times.
Then, we report the averaged value across five experiments with its 95\% confidence interval.

For crosslingual analysis, we use the Multilingual Spoken Words dataset (MSW) \cite{mazumder2021multilingual}, which contains 1-second sliced words based on the CommonVoice dataset \cite{DBLP:conf/lrec/ArdilaBDKMHMSTW20}.
However, as OMW and Epitran do not cover all languages, we choose the intersection between the supported languages: English, Chinese, Italian, Spanish, Indonesian, Polish, and Swedish.
Similarly for OMW, we randomly choose 2K word utterances per language, a total of 14K utterances, and use bootstrapping with five repeats.

\subsection{Sampling word pairs}\label{subsec:wordpairs}
In addition to {\bf Near homophone} and {\bf Synonym} pairs described in \Cref{subsec:syn_hom}, we sample three more types of word pairs for comparison.
(i) {\bf Random} pairs are generated by pairing the sampled list of words with itself after shuffling, used as a lower bound for similarity.
(ii) {\bf Same word} pairs are generated using all the unique words in the sampled set and pairing those with instances of the same word, used as an upper bound for similarity.
(iii) {\bf Same speaker} pairs are generated by pairing unique words of the same speaker.

For crosslingual analysis, we exclude the \textbf{Same speaker} condition, as OMW does not offer speaker information.
We define \textbf{Same word} as above, but for \textbf{Near homophone} and \textbf{Synonym} pairs, we restrict the languages in each pair.
One word is always English, and the paired other always comes from non-English languages.
The restriction exists to avoid the English-English pair.
Also, data for other languages is too small to generate a non-English-non-English pair.

\subsection{Extracting S3M representations}
We use the layer-wise features of \wavtovecB, \textit{-Large}, \hubertB, and \textit{-Large}.
To test the impact of different datasets, we additionally test \xlsr\  \cite{babu22_interspeech} and \wavlmL\  \cite{chen2022wavlm}, which have similar number of parameters as \wavtovecL\  and \hubertL.
\xlsr \ is pre-trained on speech from multiple languages, while the others are pre-trained on English only.
As all the models' pre-training losses contain \textbf{cosine similarity}, we use it as the similarity metric.

To extract word-level representations, we experiment with two types of slicing.
\textbf{Feature slicing} inputs the entire utterance into the model and slices the relevant time span from the resulting representation.
This method is commonly used in the analysis literature \cite{pasad2021layer,pasad2023comparative,pasad2023self}.
\textbf{Audio slicing} inputs only the specific word segment, in order to remove the contextual information surrounding the word \cite{choi2023understanding,choi2022opening}.

We also test multiple types of pooling to obtain a single vector representation for each word segment.
\textbf{Mean pooling} across the temporal dimension is the most common approach \cite{yang21c_interspeech,pasad2021layer}.
\textbf{Center pooling} chooses the temporally middle frame, which has been previously observed to retain word identity information \cite{pasad2023self}.
\textbf{Centroid pooling} measures the cosine similarity between all the frames and returns one specific frame that is most similar to others.

\section{Findings}\label{sec:analysis}
In this section, we analyze the S3M representations by comparing the cosine similarities of various word pairs.
First, we study the effect of different slicing and pooling techniques (\Cref{sec:res1,sec:res2}).
Next, we perform a comparative study of different S3Ms in our English (\Cref{sec:res3}) and crosslingual (\Cref{sec:res4}) word pairs.
Finally, we study the effect of speaker variability in our datasets (\Cref{sec:res5}). 
We provide all the code for reproducibility.\footnote{\url{https://github.com/juice500ml/phonetic_semantic_probing}}

\subsection{Feature slicing squashes representations}\label{sec:res1}
\begin{figure}[h]
    \centering
    \includegraphics[width=0.45\textwidth]{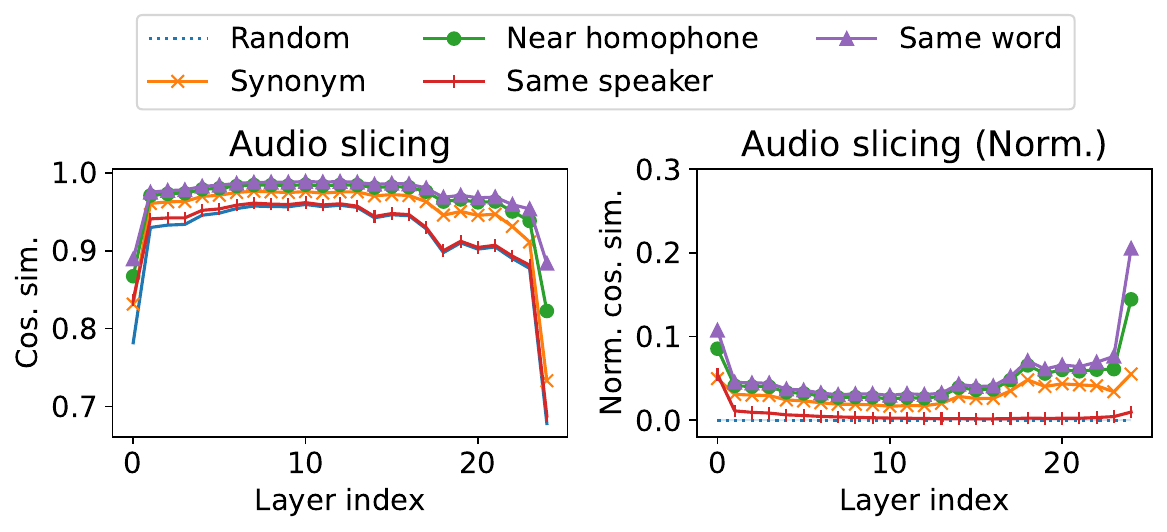}
    \includegraphics[width=0.45\textwidth]{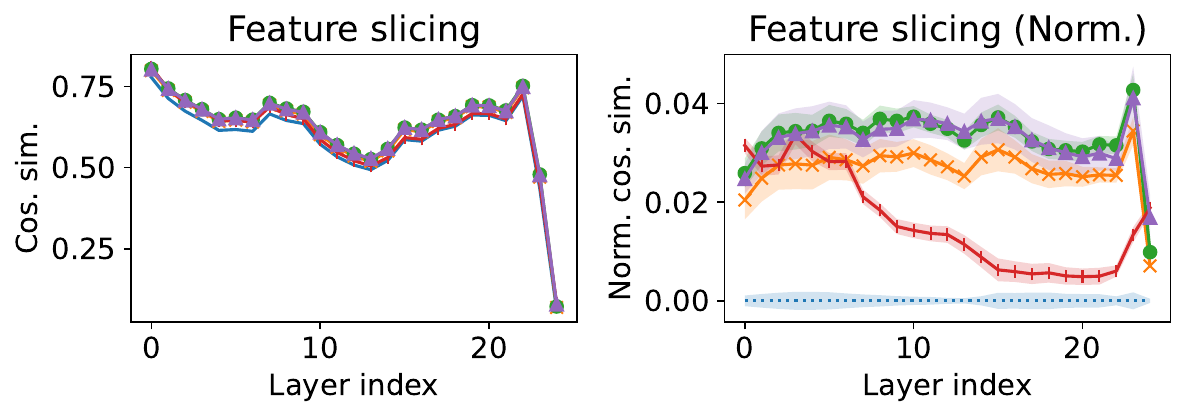}
    \caption{
        Cosine similarity between paired word representations of HuBERT-large on the
        LibriSpeech dev-clean and test-clean subsets.
        We compare audio slicing (top) vs.~feature slicing (bottom).  Normalized similarity curves (subtracting the Random baseline, right) are included to visualize the differences more clearly.
    }
    \label{fig:hubert-large}
\end{figure} 

We first compare audio and feature slicing in \Cref{fig:hubert-large}.
Each figure shows the average cosine similarity of five types of word pairs, as described in \Cref{subsec:wordpairs}.
Comparing the unnormalized plots in \Cref{fig:hubert-large}, we can immediately observe that feature slicing, unlike audio slicing, makes absolute values of similarities similar for all the word pairs, \textit{i.e.}, squashing the representations.
We suspect it is due to the transformer architecture mixing the surrounding contextual information of the whole utterance so that the resulting representations become similar.
Considering that previous work successfully extracted word information from the feature-sliced representations \cite{pasad2023self}, we emphasize that the existence of information is not equivalent to how representations are distanced from each other.
Even if the representations are squashed, the information is likely stored within the representation in a nontrivial manner, requiring additional learnable modules to extract the information.

Excluding the case where the confidence intervals overlap, the word similarities are ordered \homophone\ (green circle) $>$ \syn\ (orange X) $>$ \rand\ (blue dotted line).
This finding implies that the distance between the HuBERT-large representations encodes more phonetics than semantics across all the layers.
In order to exclusively focus on lexical semantics, we conduct the remaining analyses with audio slicing.
The audio-sliced representations do not differentiate between word senses (multiple meanings of a single word) that require contextual information.

Additionally, \Cref{fig:hubert-large} underlines the importance of having a baseline cosine similarity.
Although the similarity values are close to $1.0$, their high values are due to the anisotropic nature of transformer embeddings \cite{DBLP:conf/emnlp/Ethayarajh19}, \textit{i.e.}, representations crowded in a specific direction.
From now on, we will only report scores normalized by subtracting \rand \ (denoted Norm.) as they present more interpretable layer-wise trends while preserving relative comparisons between different types of word pairs.

\subsection{Effects of different pooling methods}
\label{sec:res2}
\begin{figure}[h]
    \centering
    \includegraphics[width=0.45\textwidth]{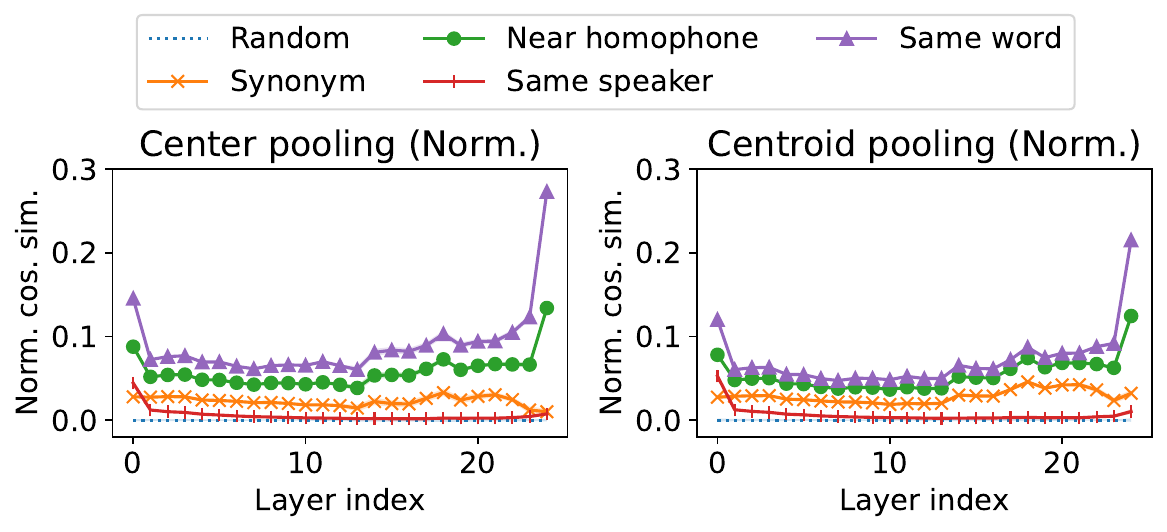}
    \caption{
        Different pooling methods for HuBERT-large representations.
    } \label{fig:hubert-large-pool}
   
\end{figure}

We compare normalized similarities of different pooling methods: mean pooling (\Cref{fig:hubert-large}), center pooling, and centroid pooling (\Cref{fig:hubert-large-pool}).
We observe that (i) centroid pooling and mean pooling retain more semantic content (\syn) than center pooling, whereas (ii) center pooling preserves more word identity information (\same) than both centroid pooling and mean pooling.
These differences suggest that the temporal center (center pooling) is not necessarily at the center of the representation space (centroid pooling).
Also, each framewise representation within a single word may model different aspects of speech.
These implications urge future work on the intra-differences of word representations.
We use the mean pooling for the remainder of our analysis.

\subsection{S3Ms encode more phonetic than semantic content}
\label{sec:res3}
\begin{figure}[h]
    \centering
    \includegraphics[width=0.45\textwidth]{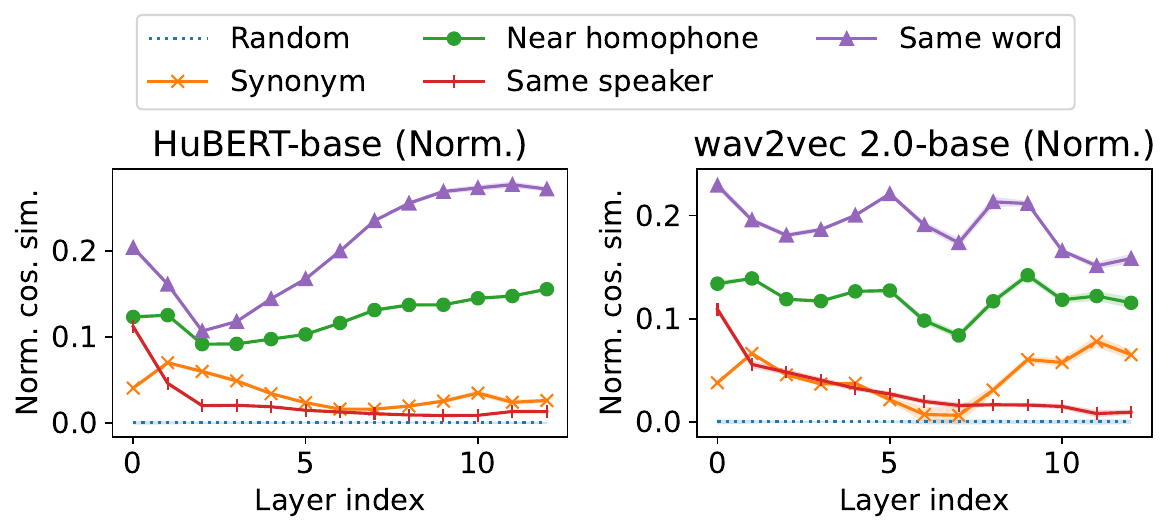}
    \includegraphics[width=0.45\textwidth]{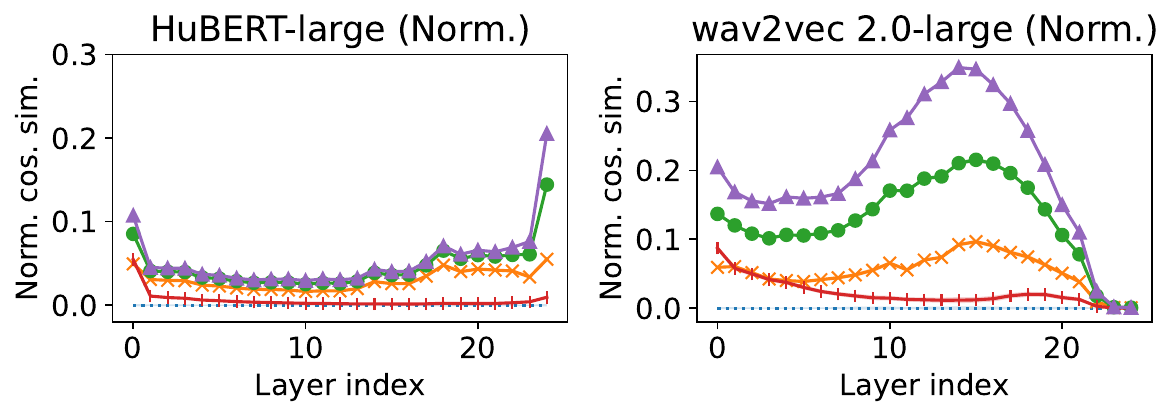}
    \includegraphics[width=0.45\textwidth]{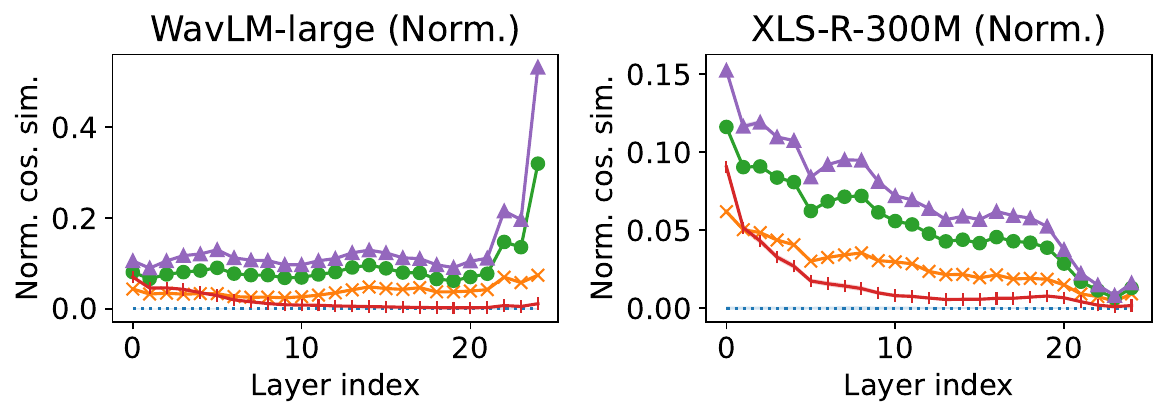}
    \caption{
        Similarities between word representations of various models.
    }
    \label{fig:other_models}
\end{figure}

In \Cref{fig:other_models}, we observe a clear dominance of \homophone\ similarity over \syn \ similarity in all the layers and all the models.
This indicates that S3Ms encode phonetically similar word pairs more closely than semantically similar pairs.
Additionally, the layer-wise trends are similar for \syn, \homophone, and \same.
This finding is corroborated by previous work studying word identity, phonetics, syntactic, and semantic content within S3M layers \cite{pasad2023self}.
Unlike pre-trained text models where the linguistic properties follow a hierarchical structure \cite{kenton2019bert}, the same S3M layers are best at encoding both phonetic and semantic linguistic properties.
Additionally, we see that the speaker information diminishes for the later layers of all S3Ms. 
Also, the layer-wise trends align with the previous findings~\cite{pasad2023comparative, sanabria2023analyzing, pasad2023self} that \hubertL\ and \wavlmL\  maintain phonetic information until the last layer, whereas \wavtovecL\  encodes most information in the middle.

\subsection{Crosslingual word pairs follow similar trends}
\label{sec:res4}
\begin{figure}[h]
    \centering
    \includegraphics[width=0.45\textwidth]{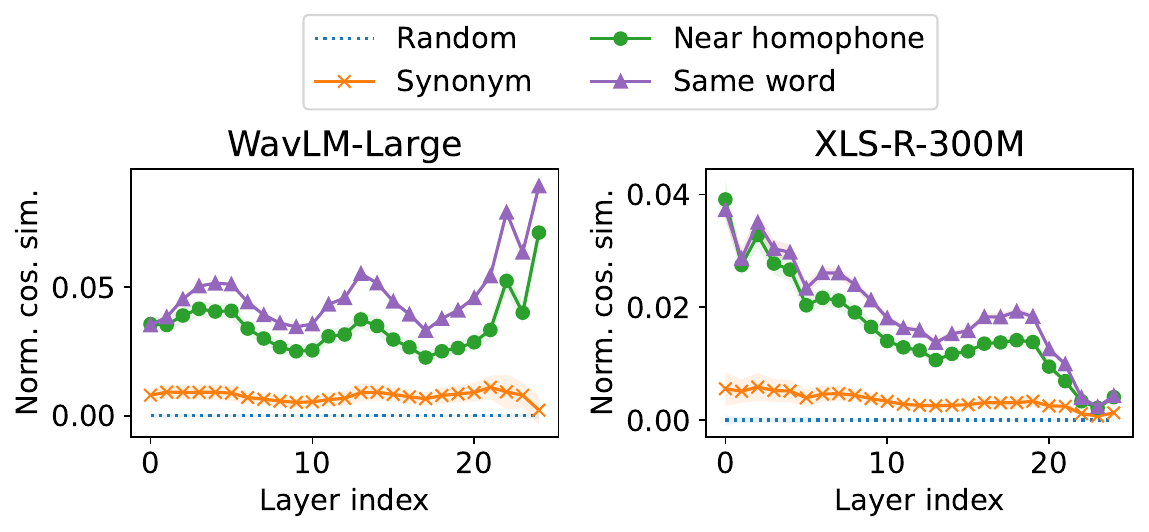}
    \caption{
        Similarities between crosslingual word representations of \wavlmL\  and \xlsr\  on MSW.
    }
    \label{fig:xling}
\end{figure}

We conduct a similar analysis using the MSW dataset in \Cref{fig:xling} and found a similar trend to the monolingual LibriSpeech dataset in \Cref{fig:other_models}.
Surprisingly, there is a non-negligible similarity between synonyms across different languages, both on cross-lingually trained \xlsr\  and English-only \wavlmL.
Nevertheless, similarly to previous observations, \syn\  pairs are consistently much farther than \homophone\ pairs.
As the paired words are always in different languages, the above results suggest that S3Ms could possess limited translation abilities.

\subsection{Single speaker utterances also follow similar trends}
\label{sec:res5}
\begin{figure}[h]
    \centering
    \includegraphics[width=0.45\textwidth]{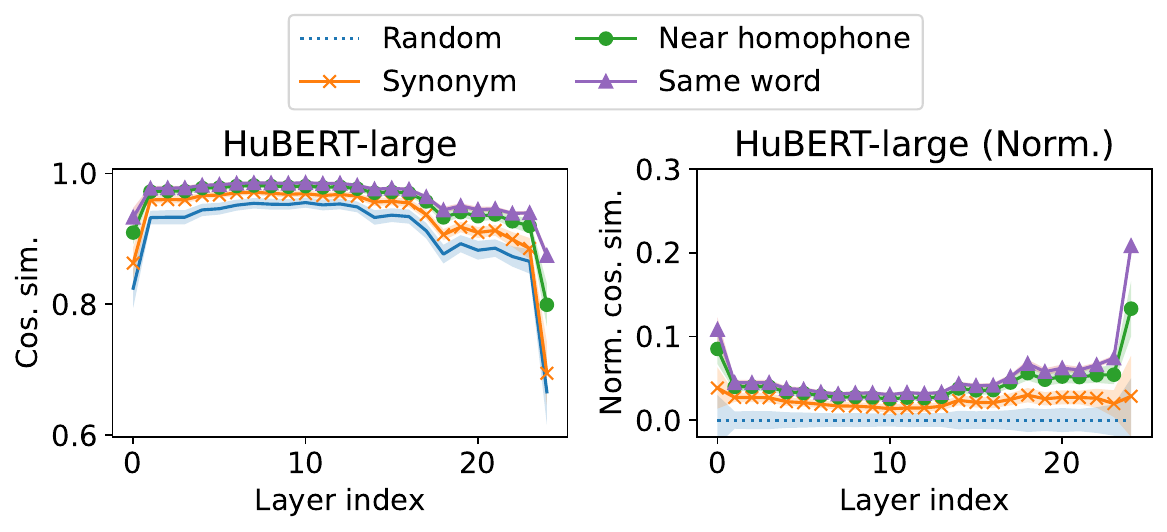}
    \caption{
        HuBERT representations of five speakers' utterances.
    } \label{fig:hubert-large-spk}
   
\end{figure}

To observe whether speaker variability affects the analysis, we consider the five speakers with the most utterances: speaker ID = (5142, 2412, 6313, 1580, 2277).
We use all samples from each speaker and calculate the similarity score per speaker.
In \Cref{fig:hubert-large-spk}, we report the mean and 95\% confidence intervals for similarity scores across five speakers.
Comparing \Cref{fig:hubert-large,fig:hubert-large-spk}, we observe that the results are nearly identical, although the variance is higher for \rand\  and \syn\  word pairs.

\section{Is Word Identity Sufficient for Intent Classification?}\label{sec:baseline}
As shown in \Cref{sec:analysis}, S3M representations encode phonetics more than semantics in their distances.
However, the downstream task performance on semantic tasks such as intent classification \cite{qian2021speech,saade2019spoken,coucke2018snips} is known to have extremely high accuracies for certain datasets \cite{yang21c_interspeech,arora21_interspeech,lugosch19_interspeech}.
Based on our findings, we question whether the word identity itself could suffice for these tasks.
To test this hypothesis, we represent each utterance as a bag of words (BoW), thus stripping away any word meaning information.
For example, the utterances ``blue sea is blue'' and ``red sea'' become $[2, 1, 1, 0]$ and $[0, 1, 0, 1]$.
Then, we compare the performance of S3Ms with BoW to verify whether S3Ms exceed our word identity baseline.

\subsection{Settings}
The \textbf{intent classification} (IC) task takes the speech utterance as input and classifies it according to the speaker's intent.
For example, the Fluent Speech Commands (FSC) dataset contains 30K utterances with 31 unique intents, such as \texttt{action:change language}, \texttt{object:newspaper}, or \texttt{location:bathroom}.
Similarly, the Snips Smartlights (SNIPS)~\cite{saade2019spoken,coucke2018snips} dataset has 1660 utterances with 6 intents, such as \texttt{SwitchLightOn} or \texttt{SetLightColor}.

We train a decision tree for BoW and a single fully-connected layer for S3Ms.
We use the original train/val/test splits and compare the test accuracy.
Additionally, we use a more challenging split, introduced by \cite{arora21_interspeech}.
Speaker split (SPK) removes the overlap of speakers between the splits, and utterance split (UTT) minimizes the overlap of n-grams.

\subsection{Results}
\begin{figure}[t]
    \centering
    \subfloat[SNIPS (Close field)]{\includegraphics[width=0.46\textwidth]{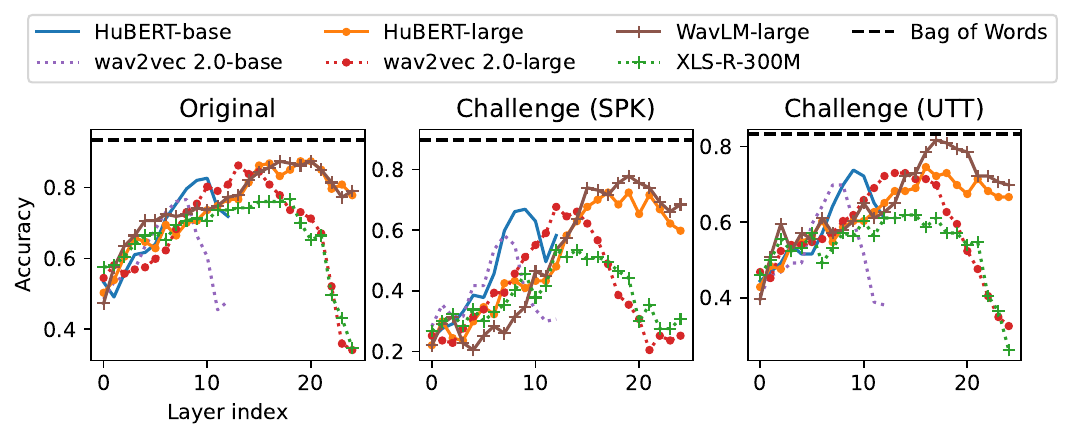}\label{fig:snips}}\\
    \subfloat[Fluent Speech Commands]{\includegraphics[width=0.46\textwidth]{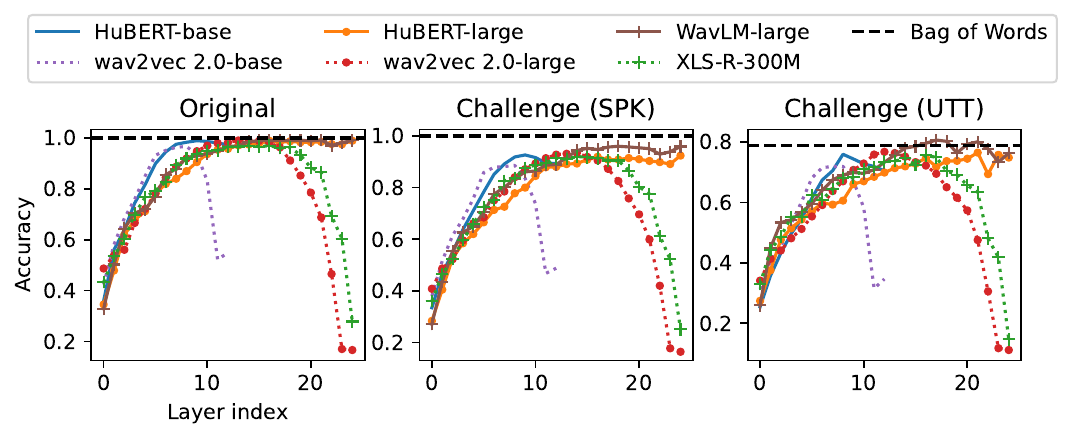} \label{fig:fsc}}
    \vspace{-0.5em}
    \caption{
        Comparing our Bag-of-Words (BoW) baseline with S3Ms on Intent Classification (IC) tasks.
        Exceeding the BoW baseline implies that S3Ms contain semantic information.
    } \label{fig:intent_cls}
\end{figure}

In \Cref{fig:intent_cls}, we see that BoW almost always outperforms or is comparable to S3Ms, suggesting that high accuracy on these tasks \cite{yang21c_interspeech} cannot conclusively indicate the semantic capabilities of S3Ms.
Moreover, BoW achieves 100\% accuracy on the original and SPK splits of FSC, implying that these evaluation settings are inadequate for measuring semantic capability.
Additionally, \wavlmL~(layer index 21) shows the best performance amongst S3Ms, aligning with previous results on word label similarity \cite{pasad2023comparative} and LibriSpeech ASR performance \cite{chang2023exploring}.

\section{Related Work}
Comparative studies on phonetic and semantic information were previously conducted on non-S3M speech models \cite{chen2023reality,chen2018phonetic,nguyen2020zero}.
We primarily focus on analyzing and comparing the layer-wise representations of S3Ms.
Recent literature on layer-wise analysis often focuses on measuring the downstream task performance using probing \cite{choi2023understanding,shen23_interspeech}, such as phoneme classification for phonology \cite{choi2023understanding}.
Canonical correlation analysis (CCA) is also often used \cite{pasad2021layer,pasad2023comparative,pasad2023self} to measure correlation with linguistically meaningful information, such as word identity \cite{pasad2023comparative,pasad2023self}.
However, both use learnable modules to verify whether the linguistic information is encoded rather than directly observing the representations via distances.
Finally, frozen S3M representations are widely used after k-means clustering \cite{wells22_interspeech,abdullah23_interspeech,chang2023exploring}.
The cluster indices for each frame are now sometimes called semantic tokens \cite{borsos2023audiolm}.
As the distances between representations determine the clusters, we believe that analyses like ours can provide further insight when using discrete tokens derived from S3Ms.

\section{Conclusion}
We analyzed various S3Ms using synonyms and near-homophone word pairs.
We conclude that phonetics dominates semantics within all layers of S3Ms.
Hence, we are further motivated to look in detail at existing semantic benchmarks for S3Ms.
We introduce a bag of words baseline that fully preserves the word identity but removes the word meaning.
We show that existing intent classification baselines can be improved by having an additional baseline.

\ifinterspeechfinal
\section{Acknowledgements}
This work is based on the results obtained from the project, Programs for Bridging the gap between R\&D and the IDeal society (society 5.0) and Generating Economic and social value (BRIDGE)/Practical Global Research in the AI × Robotics Services, implemented by the Cabinet Office, Government of Japan.
This work used the Bridges2 system at PSC and Delta system at NCSA through allocation CIS210014 from the Advanced Cyberinfrastructure Coordination Ecosystem: Services \& Support (ACCESS) program, which is supported by National Science Foundation grants \#2138259, \#2138286, \#2138307, \#2137603, and \#2138296.
\fi

\bibliographystyle{IEEEtran}
\bibliography{mybib}

\end{document}